\documentclass[10pt,twocolumn,letterpaper]{article} 

\usepackage{avss}
\usepackage{times}
\usepackage{epsfig}
\usepackage{graphicx}
\usepackage{amsmath}
\usepackage{amssymb}

\usepackage[pagebackref=true,breaklinks=true,letterpaper=true,colorlinks,bookmarks=false]{hyperref}

\avssfinalcopy 

\def\avssPaperID{****} 

\ifavssfinal\fi
\begin{document}

\title{Multi-target Multi-camera Vehicle Tracking Using Transformer-based Camera Link Model and Spatial-Temporal Information}

\author{Hsiang-Wei Huang, Cheng-Yen Yang, Jenq-Neng Hwang\\
University of Washington\\
Seattle, Washington, USA\\
{\tt\small hwhuang@uw.edu, cycyang@uw.edu, hwang@uw.edu}
}

\maketitle

\begin{abstract}
   Multi-target multi-camera tracking (MTMCT) of vehicles, which involves tracking vehicles across multiple cameras, is a crucial application for the development of smart cities and intelligent traffic systems. The main challenges of MTMCT of vehicles include the intra-class variability of the same vehicle and the inter-class similarity between different vehicles, as well as how to accurately associate the same vehicle across different cameras under a large search space. Previous methods for MTMCT typically use hierarchical clustering of trajectories to conduct cross-camera association. However, this approach does not take spatial and temporal information into consideration and the search space can be large. In this paper, we propose a transformer-based camera link model with spatial and temporal filtering to conduct cross-camera tracking. Our model achieved an IDF1 score of 73.68\% on the Nvidia Cityflow V2 dataset test set, demonstrating the effectiveness of our approach for multi-target multi-camera tracking.
\end{abstract}

\section{Introduction}
Multi-target multi-camera tracking (MTMCT) is an important technique for intelligent transportation system applications, such as traffic light and signal planning, traffic flow analysis, and journey time estimation. However, large-scale multi-camera vehicle tracking is a difficult task due to the challenges of a large search space for vehicle ReID across different cameras, intra-class variability and inter-class similarity characteristics of vehicles, and heavy occlusion that can occur during tracking. To address these issues, we abandoned the previous hierarchical method for large-scale MTMCT systems and proposed a traffic-aware transformer-based camera link model that takes spatial and temporal information into account.

Our proposed transformer-based ReID camera link model successfully reduces the search space with spatial-temporal filtering and achieves decent performance with strong appearance feature extraction.

\section{Related Work}

\subsection{Vehicle Re-ID}
Object re-identification (Re-ID) aims to associate a particular object across different camera views. Extracting reliable and discriminative features is the ultimate goal of the feature extractor model. Most recent Re-ID research uses CNN-based models as feature extractors, which have two main disadvantages compared to transformer-based models. Firstly, the smaller receptive field of CNNs, compared to transformer-based models, limits their ability to find long-range dependencies, an important component for accurate Re-ID. Secondly, the downsampling operations in CNNs can reduce the spatial resolution, resulting in information loss on the object's fine-grained features, which contain rich information for Re-ID tasks.

Recent research work utilizes transformer \cite{transreid} as feature extraction model and proposes a transformer-based model for Re-ID tasks that overcomes the limitations of CNN-based models. The transformers have a larger receptive field, which enables them to capture long-range dependencies that are critical for accurate Re-ID. Furthermore, transformers do not have downsampling operations, which helps maintain the spatial resolution of the object's features, resulting in better performance. Several transformer based method \cite{transreid} achieved state-of-the-art results on several benchmark datasets, highlighting the potential of transformer-based models for Re-ID tasks.

Due to these reasons, we incorporate transformer-based models into our MTMCT system to conduct vehicle Re-ID by using attention-mechanism to extract more discriminative features for vehicle re-identification.

\subsection{Single-camera Tracking}
Single-camera tracking (SCT) is a computer vision task that aims to locate moving objects over time in a video stream from a single camera. Recent single-camera tracking algorithms \cite{sort,deepsort,bytetrack} usually follow the tracking by detection paradigm. Firstly, an independent object detector generates detections in each frame, and then an association algorithm associates these detections to construct trajectories based on appearance features, bounding box distance \cite{huang2023observation}, or motion predicted by the Kalman filter \cite{kf}. Common tracking by detection methods include SORT \cite{sort} and DeepSORT \cite{deepsort}. These algorithms either utilize target motion as a clue for association or use both motion and appearance to construct object trajectories.

However, the tracking accuracy of SCT methods can be limited by occlusions, motion blur, and other factors. To address these issues, some recent works \cite{bytetrack,huang2023observation} introduce additional tricks into the tracking process to improve the tracking performance. These methods improve the accuracy and robustness of single-camera tracking and have potential applications in surveillance, autonomous driving, and robotics. However, the task of multi-camera tracking is still challenging due to the complexity of cross-camera association.

\subsection{Camera Link Model}
After obtaining the appearance features of vehicles, the camera link model \cite{hier1, hier3} (CLM) utilizes these features to conduct cross-camera vehicle association. Several previous works \cite{hier1, hier3, hier2} use hierarchical clustering to obtain target trajectories across cameras. However, these methods do not take spatial and temporal information into account, which can be further used to narrow down the search space of potential association candidates.

To further increase the accuracy of Re-ID, some other works \cite{2021_1, 2021_2, 2021_3} incorporate several other spatial and temporal constraints to reduce the search space, including vehicles' moving directions, transition time windows between adjacent cameras, and vehicle appearance similarity into the camera link model. By reducing the number of potential association candidates, the Re-ID accuracy can increase significantly.

Another related work that has been proposed to improve the performance of cross-camera vehicle association is the self-supervised camera link model approach \cite{hier2}. This approach considers the topology of the camera network and the spatial distribution of cameras to aid in vehicle association. Specifically, it utilizes the physical layout of the cameras to predict the likelihood of a vehicle appearing in a certain camera view based on its location in the network. This information is then used to generate a set of candidate matches for a given vehicle across multiple cameras.

\begin{figure}[t]
  \centering
  \includegraphics[width=\linewidth]{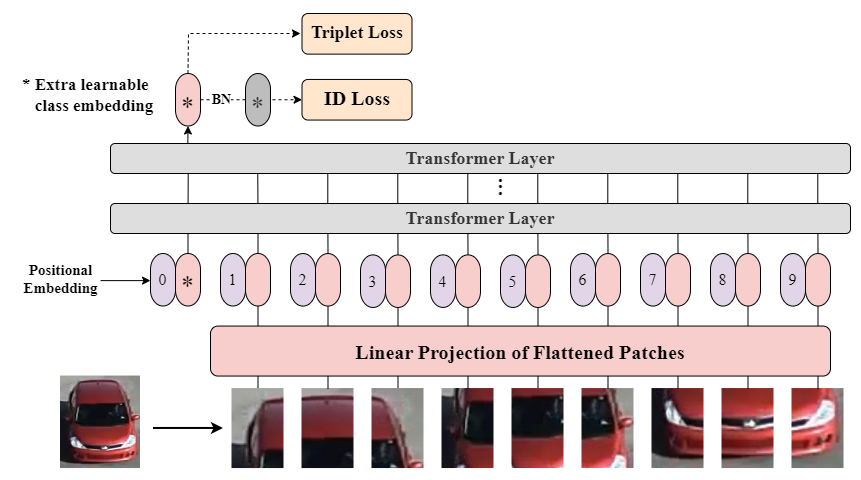}
  \caption{The framework of the transformer-based ReID model from \cite{transreid}. The BNNeck is introduced inspired by \cite{bagoftricks}.}
\end{figure}

\section{Proposed method}
The MTMCT system typically comprises three parts: single-camera tracking (SCT), vehicle Re-ID across cameras, and a camera link model that associates objects based on appearance features extracted from the Re-ID model. The lower bound performance of the MTMCT system is typically determined by the quality of SCT and the extent to which spatial-temporal information is utilized in the camera link model. If the MTMCT system can achieve good results in SCT and incorporate spatial-temporal information, it can narrow down the search space for Re-ID across cameras and achieve a certain level of accuracy in multi-camera tracking. However, due to inter-class similarity between vehicles, there may be some challenging cases where two similar cars have the same moving direction and similar transition time windows across cameras. These cases can cause confusion for the MTMCT system and result in cross-camera ID switching. To solve these challenges and achieve higher accuracy in multi-camera tracking, a strong Re-ID model capable of extracting more discriminative features from cars with similar appearances is needed. Therefore, the upper bound performance of the MTMCT system depends on whether the Re-ID model can address these challenging cases by extracting more discriminative features.

\subsection{Transformer-based Vehicle Re-ID}
Re-ID is a fundamental task in MTMCT that determines the upper bound performance of the entire system. One of the key challenges in this task is the high inter-class similarity between vehicles, which makes it difficult to extract robust features. To overcome this challenge, we exploit the effectiveness of transformer models for Re-ID. By incorporating the transformer-based baseline model TransReID \cite{transreid}, we can extract more discriminative features for both single-camera tracking and cross-camera Re-ID. To account for the large visual appearance variation between different cameras, lighting conditions, and vehicle orientations, we incorporate both triplet loss and cross-entropy loss for global features during model optimization.

The triplet loss \cite{triplet} focuses on optimizing the distance of a triplet set {\textit{a,p,n}} in the feature space by making the distance between positive pairs smaller than negative pairs by a margin \textit{m}:

\begin{equation}
    \label{eqn:triplet}
    \textit L_{triplet}
    = \sum_{i=1}^Nmax(d(f_i^a,f_i^n)-d(f_i^a,f_i^p)+m,0)
\end{equation}

\noindent where $N$ is the total number of training samples $f^a$, $f^p$, $f^n$ are the features of anchor, positive and negative samples extracted from the Re-ID model. $d(\cdot)$ is the representation of the distance between two features, which can be either calculated from cosine distance or Euclidean distance.\\
\hspace*{1em}Cross-entropy loss is also used in the network optimization. Which can be formulated as:

\begin{equation}
    \label{eqn:ce}
    \textit L_{ce}
    = \sum_{i=1}^N-q_ilog(p_i)
    \begin{cases}
    q_i = 0, y \neq i\\
    q_i = 1, y = i
    \end{cases}
\end{equation}

\noindent where y is the groundtruth ID label and $p_i$ as the prediction for class $i$.

\begin{figure}[h]
  \centering
  \includegraphics[width=\linewidth]
  {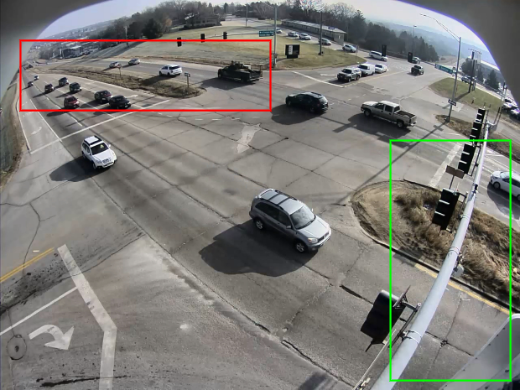}
  \caption{Vehicles detected entering the intersection while the traffic light is red will be matched with trajectories in the current camera. On the other hand, vehicles leaving the camera's field of view through the green traffic zone will be matched with trajectories in the next adjacent camera.}
\end{figure}

\subsection{Detection and Single-camera Vehicle Tracking}

Detection is a crucial factor that affects the performance of tracking. To produce high-quality detections, we incorporate Cascade-RCNN \cite{cascadercnn} for vehicle detection. First, the vehicle detection model is pretrained on the COCO dataset and then fine-tuned on the training and validation data of the AI City Challenge dataset \cite{cityflow}.

We adopt the framework of DeepSORT \cite{deepsort} with several modifications as our algorithm for single-camera tracking. DeepSORT utilizes Kalman filter to predict motion and combines appearance and motion similarity to conduct association.

To prevent false positive detections, all final trajectories with short length are filtered out and will not be associated in the camera link model.

In addition to the filtering of short trajectories, we also implement the confidence score cascade association proposed in \cite{bytetrack} to increase the association accuracy. In our pipeline, high confidence score detection is first associated, and those unmatched detection and low confidence score detection will conduct association in the second round. Finally, those unassociated detection will be initiated as new tracklets.

Overall, these modifications to the DeepSORT framework help to improve the performance and reliability of our single-camera tracking algorithm.

\subsection{Trajectory-based Camera Link Model}

After obtaining the SCT results for each camera, the camera link model (CLM) can associate trajectories between adjacent cameras based on several clues, including moving directions, transition time windows, and appearance similarity.\\
\noindent\textbf{Moving directions.} The moving direction of the vehicles is important for the camera link model to filter impossible association candidates. To obtain the moving direction, we use entry/exit traffic zones to classify the moving direction of the vehicles. The traffic zones are manually labeled to obtain the correct moving direction of each vehicle. The first zone and the last zone the car passes through will be considered as the entry and exit zone of the car, respectively. Based on the entry/exit zone information, we can classify the vehicle's moving direction.

\noindent\textbf{Bi-directional transition time windows.} The transition time window for each camera pair can be determined by the distance between two cameras, traffic signal condition, and several other factors. To ensure perfect time zones for camera link model association, the time window is fine-tuned carefully based on these factors. Note that the optimal time windows of a camera pair can be different according to the moving direction of the vehicle. This means the optimal time window for $cam.A$ \textrightarrow{} $cam.B$ and $cam.B$ \textrightarrow{} $cam.A$ can be different. This can be caused by the difference in traffic signal condition in the two moving directions and the camera's rotation angle. Thus, we proposed bi-directional transition time windows (BTT) that enable different transition time windows that take moving directions into account so that the camera link model can have the best optimal time window for cross-camera association.

\noindent\textbf{Appearance similarity.} The trajectories matching across cameras are performed with the Hungarian algorithm. To further reduce the search space of association candidates, temporal mask and direction mask (TDM) are taken into account for association between camera pairs. By introducing the temporal and direction mask, the final similarity for each trajectory pair $Traj_i$ and $Traj_j$ is defined by the following formula:

\begin{equation}
    \label{eqn:clm}
    \textit S_{ij}
    = 
    d(f_i,f_j) * T_{mask} * Dir_{mask}
\end{equation}

Here, $d(\cdot)$ is the distance between two trajectory-based features, which is an average pooling of all the features for each trajectory from the SCT result. $T_{mask}$ is the temporal mask, which represents the time window filtering mechanism in the camera link model. With $t1$ and $t2$ as the lower and upper bound of the time window and $t_{appear}$ as the appearance time of the vehicle, we define $T_{mask}$ as:

\begin{equation}
    \label{eqn:time}
    \textit T_{mask}
    = 
    \begin{cases}
    1, t_1<t_{appear}<t_2\\
    0, otherwise
    \end{cases}
\end{equation}

$Dir_{mask}$ is the direction mask, which can be derived based on the moving direction reasoning of the vehicle. It is defined as:

\begin{equation}
    \label{eqn:dir}
    \textit Dir_{mask}
    = 
    \begin{cases}
    0, direction\; conflict\\
    1, otherwise
    \end{cases}
\end{equation}

After obtaining the similarity matrix, the camera link model performs the Hungarian algorithm for each camera pair, associates cross-camera trajectories, and filters out associations with low similarity. Finally, a unique global ID is assigned to each trajectory after association.

\begin{figure}[h]
  \centering
  \includegraphics[width=\linewidth]
  {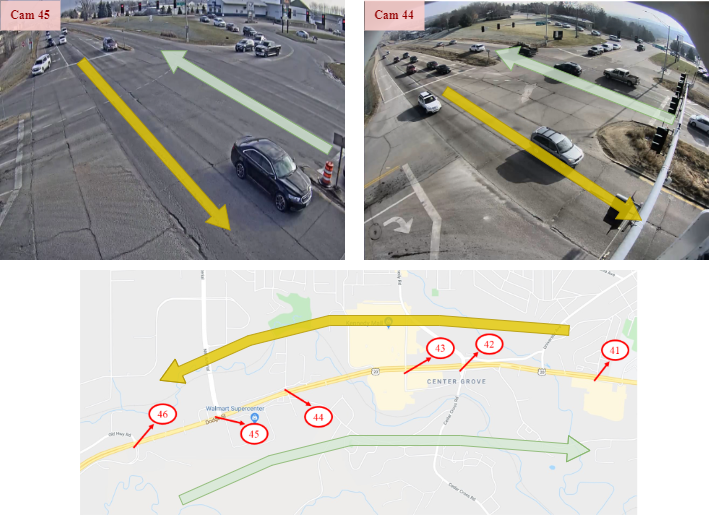}
  \caption{This picture illustrate the direction conflict situation. Vehicles going in the yellow direction will not be matched with trajectories going in the green direction in the adjacent camera. The direction can be obtained by the traffic zone.}
\end{figure}

\section{Experiments}

\subsection{Datasets}
We conducted experiments using the CityFlowV2 dataset \cite{cityflow}. The dataset consists of 46 camera views, covering a diverse set of locations such as highways, intersections, city streets, and residential areas across a mid-size U.S. city. It includes a total of 313,931 bounding boxes and 880 annotated vehicle identities. The training and validation sets contain several intersections with multi-cameras, resulting in overlapping field of views. However, in the testing set, the six cameras are separate and located at six different intersections with no overlapping field of view.

\hspace*{1em}IDF1 \cite{idf1} is adopted as the evaluation metric for the MTMCT system. Let $IDTP$ denotes the count of true positive IDs, $IDTN$ as the count of true negative IDs, $IDFP$ as the count of false positive IDs, and $IDFN$ as the count of false negative IDs. The IDF1 can be calculated with the following formula:

\begin{equation}
    \label{eqn:idf1}
    \textit IDF1
    = \dfrac{2IDTP}{2IDTP+IDFP+IDFN}
\end{equation}

\subsection{Implementation Details}
\noindent\textbf{Vehicle Re-ID.} We use stochastic gradient descent (SGD) optimizer for training with a momentum of 0.9 and weight decay of 1e-4, and an initial learning rate of 0.008. We adopt cosine weight decay during training and apply several data augmentation methods, such as color normalization, random crop, and random flip.

\noindent\textbf{Single-camera Tracking.} We set the detection score threshold to 0.1. The similarity association threshold for SCT is 0.45, and the momentum term in the exponential moving average is 0.9.

\noindent\textbf{Camera Link Model.} We set the cosine distance threshold for cross-camera association to 0.45. We filter trajectories with a length less than 2 frames or more than 2000 frames, as they might be false detections or not moving vehicles (which are not labeled in the CityFlow dataset), respectively.

\begin{table}[h]
  \begin{center}
    {\small{
\begin{tabular}{|l|c|c|c|}
\hline
Method & IDF1 & IDP & IDR\\ 
\hline
Baseline & 34.74&  29.33 & 42.61\\ 
+TDM & 57.09 (+22.35) & 53.92 (+24.59) & 60.65 (+18.04)\\
+BTT & \textbf{73.68 (+26.59)} & \textbf{67.73 (+13.81)} & \textbf{80.77 (+20.12)} \\
\hline

\end{tabular}
}}
\end{center}
\caption{The performance of different methods on the CityFlowV2 test set. TDM stands for temporal and direction mask. BTT stands for Bi-direction transition time windows.}
\end{table}
\subsection{Results}
We evaluate the performance of our MTMCT system on the CityFlowV2 test set using several metrics, including ID Precision, ID Recall, IDF1, and MOTA. The baseline method simply performs association between each camera pair without considering any spatial-temporal information, such as the vehicle's moving direction or transition time window. After incorporating the temporal and direction masks in the camera link model, the search space is largely reduced, resulting in significant improvements in IDF1 by 22.35\%. Finally, the bi-directional time window helps the CLM obtain a better search space, further improving the performance of IDF1 by 26.59\%, achieving a final score of 73.68\% in IDF1.

We submit our proposed system to the AI City Challenge evaluation system and compare it with several different teams and methods on the same testing dataset. Our algorithm outperforms all of the teams in the 2020 AI City Challenge \cite{2020_aic} and most of the teams in the 2021 AI City Challenge \cite{2021_aic} in the City-scale MTMC vehicle tracking track on the CityFlowV2 benchmark.

\begin{table}[h]
  \begin{center}
    {\small{
\begin{tabular}{|c|l|c|}
\hline
Rank & Team & IDF1 \\ 
\hline
- & \textbf{Ours} & \textbf{73.68}\\
1 & CMU& 45.85\\ 
2 & XJTU& 44.00\\
5 & UMD& 12.45\\
6 & UAlbany&6.20\\
\hline

\end{tabular}
}}
\end{center}
\caption{Comparison with the participants in 2020 AI City Challenge \cite{2020_aic} City-scale MTMC vehicle tracking leaderboard. Our performance outperform all the teams in 2020 on the same benchmark.}
\end{table}

\begin{table}[h]
  \begin{center}
    {\small{
\begin{tabular}{|c|l|c|}
\hline
Rank & Team & IDF1 \\ 
\hline
1 & Alibaba-UCAS & 80.95\\ 
2 & Baidu & 77.87\\
3 & SJTU & 76.51 \\
- & \textbf{Ours} & 73.68\\
4 & Fraunhofer& 69.10 \\
6 & Fiberhome& 57.63\\
9 & NTU&54.58\\
\hline

\end{tabular}
}}
\end{center}
\caption{Comparison with the participants in 2021 AI City Challenge \cite{2021_aic} City-scale MTMC vehicle tracking leaderboard.}
\end{table}

\section{Conclusion}
In this paper, we present a new MTMCT system for vehicle tracking based on a transformer architecture. Our proposed system combines the strengths of single-camera tracking, vehicle Re-ID, and cross-camera association in a unified framework. The system employs a deep learning-based approach to perform robust vehicle detection and tracking in each individual camera, followed by a transformer-based camera link model to conduct accurate cross-camera association. Through the use of temporal and direction masks, we were able to largely reduce the search space of potential association candidates, leading to significant improvements in the overall performance of the system.
The experimental results on the CityFlowV2 test set demonstrate that our proposed method achieves a competitive performance in terms of IDF1, with a score of 73.68\%, demonstrating the effectiveness and robustness of our approach in multi-camera tracking. 

{\small
\bibliographystyle{ieee}
\bibliography{egbib}

\begin{thebibliography}{10}\itemsep=-1pt

\bibitem{sort}
A.~Bewley, Z.~Ge, L.~Ott, F.~Ramos, and B.~Upcroft.
\newblock Simple online and realtime tracking.
\newblock In {\em 2016 IEEE international conference on image processing
  (ICIP)}, pages 3464--3468. IEEE, 2016.

\bibitem{cascadercnn}
Z.~Cai and N.~Vasconcelos.
\newblock Cascade r-cnn: Delving into high quality object detection.
\newblock In {\em Proceedings of the IEEE conference on computer vision and
  pattern recognition}, pages 6154--6162, 2018.

\bibitem{transreid}
S.~He, H.~Luo, P.~Wang, F.~Wang, H.~Li, and W.~Jiang.
\newblock Transreid: Transformer-based object re-identification.
\newblock In {\em Proceedings of the IEEE/CVF international conference on
  computer vision}, pages 15013--15022, 2021.

\bibitem{triplet}
A.~Hermans, L.~Beyer, and B.~Leibe.
\newblock In defense of the triplet loss for person re-identification.
\newblock {\em arXiv preprint arXiv:1703.07737}, 2017.

\bibitem{hier3}
H.-M. Hsu, J.~Cai, Y.~Wang, J.-N. Hwang, and K.-J. Kim.
\newblock Multi-target multi-camera tracking of vehicles using metadata-aided
  re-id and trajectory-based camera link model.
\newblock {\em IEEE Transactions on Image Processing}, 30:5198--5210, 2021.

\bibitem{hier2}
H.-M. Hsu, Y.~Wang, J.~Cai, and J.-N. Hwang.
\newblock Multi-target multi-camera tracking of vehicles by graph auto-encoder
  and self-supervised camera link model.
\newblock In {\em Proceedings of the IEEE/CVF Winter Conference on Applications
  of Computer Vision}, pages 489--499, 2022.

\bibitem{hier1}
H.-M. Hsu, Y.~Wang, and J.-N. Hwang.
\newblock Traffic-aware multi-camera tracking of vehicles based on reid and
  camera link model.
\newblock In {\em Proceedings of the 28th ACM International Conference on
  Multimedia}, pages 964--972, 2020.

\bibitem{huang2023observation}
H.-W. Huang, C.-Y. Yang, S.~Ramkumar, C.-I. Huang, J.-N. Hwang, P.-K. Kim,
  K.~Lee, and K.~Kim.
\newblock Observation centric and central distance recovery for athlete
  tracking.
\newblock In {\em Proceedings of the IEEE/CVF Winter Conference on Applications
  of Computer Vision}, pages 454--460, 2023.

\bibitem{kf}
R.~E. Kalman.
\newblock A new approach to linear filtering and prediction problems, 1960.
\newblock J. Fluids Eng., 82(1):35–45.

\bibitem{2021_1}
C.~Liu, Y.~Zhang, H.~Luo, J.~Tang, W.~Chen, X.~Xu, F.~Wang, H.~Li, and Y.-D.
  Shen.
\newblock City-scale multi-camera vehicle tracking guided by crossroad zones.
\newblock In {\em Proceedings of the IEEE/CVF Conference on Computer Vision and
  Pattern Recognition}, pages 4129--4137, 2021.

\bibitem{bagoftricks}
H.~Luo, Y.~Gu, X.~Liao, S.~Lai, and W.~Jiang.
\newblock Bag of tricks and a strong baseline for deep person
  re-identification.
\newblock In {\em Proceedings of the IEEE/CVF conference on computer vision and
  pattern recognition workshops}, pages 0--0, 2019.

\bibitem{2020_aic}
M.~Naphade, S.~Wang, D.~Anastasiu, Z.~Tang, M.-C. Chang, X.~Yang, L.~Zheng,
  A.~Sharma, R.~Chellappa, and P.~Chakraborty.
\newblock The 4th ai city challenge, 2020.

\bibitem{2021_aic}
M.~Naphade, S.~Wang, D.~C. Anastasiu, Z.~Tang, M.-C. Chang, X.~Yang, Y.~Yao,
  L.~Zheng, P.~Chakraborty, C.~E. Lopez, et~al.
\newblock The 5th ai city challenge.
\newblock In {\em Proceedings of the IEEE/CVF Conference on Computer Vision and
  Pattern Recognition}, pages 4263--4273, 2021.

\bibitem{idf1}
E.~Ristani, F.~Solera, R.~Zou, R.~Cucchiara, and C.~Tomasi.
\newblock Performance measures and a data set for multi-target, multi-camera
  tracking.
\newblock In {\em European conference on computer vision}, pages 17--35.
  Springer, 2016.

\bibitem{cityflow}
Z.~Tang, M.~Naphade, M.-Y. Liu, X.~Yang, S.~Birchfield, S.~Wang, R.~Kumar,
  D.~Anastasiu, and J.-N. Hwang.
\newblock Cityflow: A city-scale benchmark for multi-target multi-camera
  vehicle tracking and re-identification.
\newblock In {\em Proceedings of the IEEE/CVF Conference on Computer Vision and
  Pattern Recognition}, pages 8797--8806, 2019.

\bibitem{deepsort}
N.~Wojke, A.~Bewley, and D.~Paulus.
\newblock Simple online and realtime tracking with a deep association metric.
\newblock In {\em 2017 IEEE international conference on image processing
  (ICIP)}, pages 3645--3649. IEEE, 2017.

\bibitem{2021_3}
M.~Wu, Y.~Qian, C.~Wang, and M.~Yang.
\newblock A multi-camera vehicle tracking system based on city-scale vehicle
  re-id and spatial-temporal information.
\newblock In {\em Proceedings of the IEEE/CVF Conference on Computer Vision and
  Pattern Recognition}, pages 4077--4086, 2021.

\bibitem{2021_2}
J.~Ye, X.~Yang, S.~Kang, Y.~He, W.~Zhang, L.~Huang, M.~Jiang, W.~Zhang, Y.~Shi,
  M.~Xia, et~al.
\newblock A robust mtmc tracking system for ai-city challenge 2021.
\newblock In {\em Proceedings of the IEEE/CVF Conference on Computer Vision and
  Pattern Recognition}, pages 4044--4053, 2021.

\bibitem{bytetrack}
Y.~Zhang, P.~Sun, Y.~Jiang, D.~Yu, F.~Weng, Z.~Yuan, P.~Luo, W.~Liu, and
  X.~Wang.
\newblock Bytetrack: Multi-object tracking by associating every detection box.
\newblock In {\em European Conference on Computer Vision}, pages 1--21.
  Springer, 2022.

\end{thebibliography}
}


\end{document}